%% file: main.tex
\documentclass{article} 
\usepackage{iclr2027_conference,times}

\input{math_commands.tex}

\usepackage{hyperref}
\usepackage{url}
\usepackage{graphicx}
\usepackage{booktabs}
\usepackage[table]{xcolor}
\usepackage{makecell}
\usepackage{multirow}
\usepackage{color}
\usepackage{amssymb}
\usepackage{tabularx}
\usepackage{array}

\newcommand{\name}[0]{\textsc{Self-Saliency}}
\newcommand{\namemean}[0]{\name{}$_{\mathrm{mean}}$}

\title{Look Where You Say You're Looking: Self-Grounded Attention for Visual Reasoning}

\author{
Uri Berger$^{1,2,3}$ \qquad
Gal Chechik$^{1,4}$ \qquad
Gal Dalal$^{1}$ \\[3pt]
$^1$NVIDIA Research \\
$^2$The Hebrew University of Jerusalem \\
$^3$University of Melbourne \\
$^4$Bar-Ilan University \\
\texttt{uri.berger2@mail.huji.ac.il}
}

\iclrfinalcopy 
\begin{document}

\maketitle
\lhead{} 

\begin{abstract}
\input{sections/00_abstract}
\end{abstract}

\section{Introduction}
\input{sections/01_introduction}

\section{Related Work}
\label{sec:related_work}
\input{sections/02_related_work}

\section{Method}
\label{sec:method}
\input{sections/03_method}

\section{Experiments}
\label{sec:experiments}
\input{sections/04_experiments}

\section{Attention Analysis and Ablations} \label{sec:analysis}
\input{sections/05_attention_analysis}

\section{Conclusion}
\input{sections/06_conclusion}

\subsection*{AI use statement}

We used generative AI tools to assist with code generation and to improve the wording of text initially written by the authors. We also used AI tools to support literature discovery, while the literature review itself was conducted primarily through standard scholarly search and alerting tools. In addition, we used generative AI to create synthetic data for training the reasoning-step classifier described in Appendix~\ref{app:steps_classifier}.

Generative AI was not used to generate paper sections, develop the core ideas of the work, or determine research directions. The authors take full responsibility for all content in the paper, including any text, claims, code, or other artifacts produced with the assistance of generative AI tools.

\subsection*{Ethics statement}

This work does not involve the collection of new data or interactions with human participants. All experiments are conducted using existing publicly available datasets, which we use in accordance with their respective licenses and terms of use. We do not introduce or release new datasets containing personal or sensitive information. Accordingly, we do not identify additional ethical concerns beyond those already associated with the underlying datasets and models used in this work.





\bibliography{iclr2027_conference}
\bibliographystyle{iclr2027_conference}

\appendix
\section{Further Implementation Details} \label{app:implementation_details}
\input{sections/appendix_a}

\section{Extended Results} \label{app:extended_results}
\input{sections/appendix_b}

\section{Experimenting With A Different Saliency Scoring Function} \label{app:md_baseline}
\input{sections/appendix_c}

\section{Reimplementing Baselines Without a Published Checkpoint} \label{app:reimplementing}
\input{sections/appendix_d}

\section{Attention Analysis Model Details} \label{app:model_details}
\input{sections/appendix_e}

\end{document}

%% file: math_commands.tex
\usepackage{amsmath,amsfonts,bm}

\def\eqref#1{equation~\ref{#1}}

\def\1{\bm{1}}

\DeclareMathAlphabet{\mathsfit}{\encodingdefault}{\sfdefault}{m}{sl}
\SetMathAlphabet{\mathsfit}{bold}{\encodingdefault}{\sfdefault}{bx}{n}

\newcommand{\R}{\mathbb{R}}



%% file: sections/00_abstract.tex
We introduce \name{}, a method for training Vision-Language Models (VLMs) to increase the alignment between their visual attention and the image regions mentioned in their reasoning. \name{} uses a grounding model to localize the objects mentioned in each reasoning step and treats the resulting areas as supervision for the model's visual attention.
Previous work on steering visual attention determines target image regions based solely on the image and question. In contrast, we show that conditioning the target regions on the model's generated reasoning improves downstream performance. For proper evaluation, we build a unified, broad suite of 25 visual reasoning benchmarks, where we reproduce the results of previous methods. We find that \name{} significantly outperforms both prior attention-steering methods and baselines that ground image-level text, achieving both a better average rank, and better mean score. Post-training analysis shows that the model primarily adapts its reasoning text to existing attention patterns, producing shorter steps that refer to larger regions. Nevertheless, when controlling for generated text, attention to grounded regions increases significantly across the relevant layer. 
Finally, we identify a consistent geometric bias in VLM visual attention toward the image border. However, our ablations show that \name{}'s gains cannot be explained by simply aligning attention with the center of the image, highlighting the importance of aligning visual attention with the regions mentioned in the model's reasoning.

%% file: sections/01_introduction.tex
\begin{figure} [tb]
    \centering
    \includegraphics[width=0.8\textwidth]{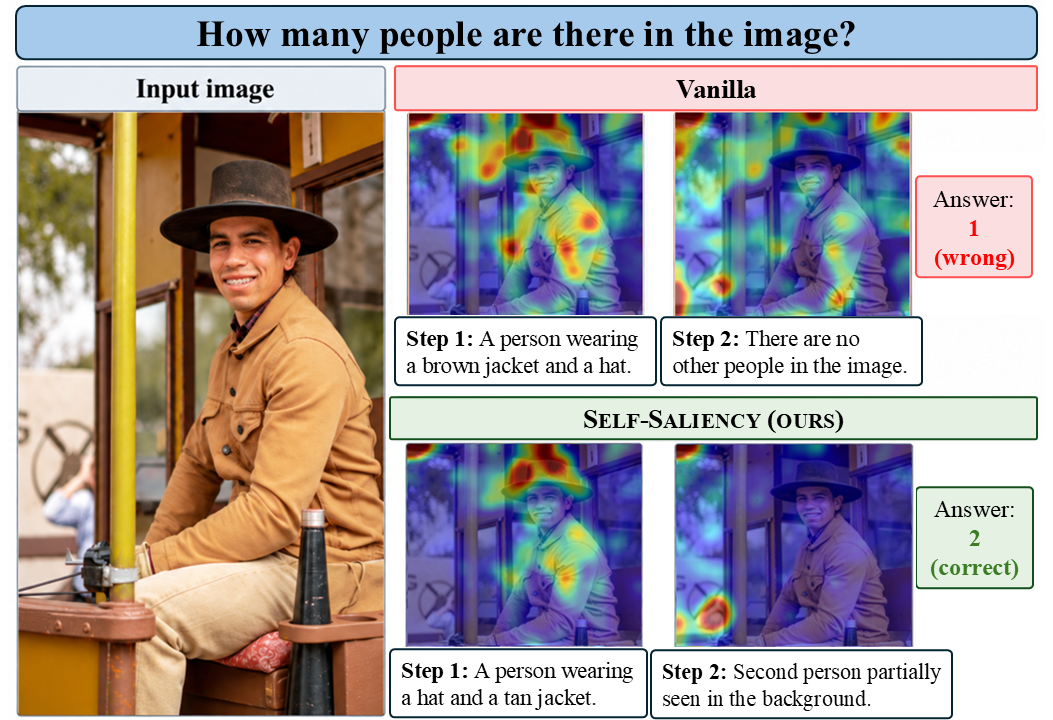}
    \caption{\name{} is trained to increase the alignment between visual attention and the regions referenced in its reasoning chain, creating two complementary pressures: (1) generate claims that can be grounded in the image, and (2) attend to the regions those claims refer to.}
    \label{fig:figure1}
\end{figure}

When answering questions about images, humans shift their gaze across different image regions. Vision-language models (VLMs) exhibit an analogous mechanism during visual reasoning: generated text tokens incorporate visual information by attending to image tokens. Surprisingly, however, the regions emphasized by VLMs differ substantially from those attended to by humans.
Indeed, a
growing body of work shows that visual attention often fails to focus on image regions that humans consider important for answering the question~\citep{gong2024damro, darcet2024vision, zhang2025beyond}.
Numerous   methods attempted to address this mismatch by post-training steering~\citep{gong2026saliency, hu2026attend}.
All these methods determine the target salient regions solely from the input image and question. By prescribing where the model should attend, they also constrain the reasoning paths available to the model. However, there is no reason to assume that these prescribed paths are optimal, or that other reasoning paths could not also lead to the correct answer.
This motivates an unexplored alternative: deriving the target regions from the model’s own reasoning chain.

To this end, we propose a post-training method, \name{} (Figure~\ref{fig:figure1}). During training, \name{} uses a grounding model to localize the image regions that are referenced in each reasoning step. The trained model is rewarded for greater alignment between its visual attention and these grounded regions.
 At inference time, the trained model doesn't utilize external models.

Inspecting recent literature, we find no broad consensus on evaluation benchmarks for visual reasoning. A large number of benchmarks have been introduced in recent years, with individual papers typically reporting results on only a subset of them. To provide a more comprehensive evaluation, we identify the benchmarks used in recent work and construct a suite of 25 benchmarks.

We evaluate \name{} on this broad suite and find that it substantially outperforms relevant baselines, with particularly strong gains on tasks that require attending to specific image regions. Surprisingly, these improvements are not limited to natural images. Although the grounding model was trained primarily on natural-image data, \name{} also improves performance on visual mathematical reasoning tasks.

We analyze whether training primarily aligns the generated text with existing attention patterns or shifts attention toward the generated reasoning. \name{} produces shorter reasoning steps that refer to larger image regions. To isolate attention changes from this shift in text, we teacher-force the pre- and post-training models on the same reasoning chains. While grounded-region attention does not increase in the two heads used for the saliency reward, it increases significantly when averaged across all heads in the relevant layer.

Finally, we analyze visual attention patterns across several VLM architectures and identify a consistent geometric bias: VLMs allocate disproportionately high attention to regions near the image border, whereas human annotations of task-relevant regions are more concentrated toward the image center. Motivated by this observation, we demonstrate that simply steering attention toward the image center does not improve downstream performance over the baselines, indicating that \name{}'s gains cannot be explained by a generic center bias. Finally, \name{} outperforms a baseline that grounds image-level text, suggesting that grounding the model's reasoning chain provides a more effective training signal.









%% file: sections/02_related_work.tex
\paragraph{Reasoning VLMs.} Recent work has increasingly focused on training VLMs to perform explicit reasoning over images. Several studies investigate the construction of multimodal chain-of-thought data for this purpose~\citep{yang2025r1, huang2026vision}.
Others modify the training procedure itself to improve reasoning behavior~\citep{zhang2025r1, wang2026vl}.

\paragraph{Training VLMs to relocate attention.}
Several previous studies have proposed methods for redistributing visual attention.
VAR~\citep{kang2025see} and LASER~\citep{yuan2026laser} transfer attention mass from sink tokens to non-sink tokens.
VGA~\citep{zhao2026tell} operates at inference time, steering attention toward image patches whose visual representations are more strongly associated with the objects referenced in the question.
RAL~\citep{li2026reinforced} encourages the model's attention distribution to move toward attention patterns of correct reasoning chains.

Most closely related to our approach, several prior works steer attention toward content-dependent ground-truth regions. KLAL~\citep{esmaeilkhani2026direct} and Saliency-R1~\citep{gong2026saliency} use human-annotated bounding boxes as supervision targets. EASE~\citep{hu2026attend} prompts GPT to identify phrases that are important for answering the question, grounds these phrases using a grounding model, and treats the resulting regions as ground truth. To the best of our knowledge, \name{} is the first method to derive grounded supervision directly from the reasoning chain generated by the policy being trained.

\paragraph{Visual attention analysis.}
In recent years, several studies have investigated how visual attention is distributed across an image. \citet{darcet2024vision} and \citet{gong2024damro} argue that a large fraction of visual attention is assigned to outlier tokens with large norms. \citet{kang2025see} define \emph{visual sink tokens} as tokens that consistently receive high attention due to their intrinsic properties, rather than their relevance to the generated text. They further suggest that such sink behavior arises from large activations in specific dimensions of the tokens' hidden states. Other studies show that LLaVA~\citep{liu2024improved} allocates a disproportionate amount of attention to the lower part of the image~\citep{zhang2025beyond,zhao2025mca}. In Section~\ref{sec:analysis}, we extend this line of analysis to several VLMs and show that visual attention is consistently concentrated along the geometric border of the image.

%% file: sections/03_method.tex
\subsection{Preliminaries}
\noindent\textbf{Image Patches.} \label{sec:patches}
Most modern VLMs begin image processing by dividing the image into a grid of disjoint patches. A vision transformer jointly processes the patches and outputs a contextualized representation.
These visual representations are then passed through an adapter that maps them into visual tokens. Thus, each visual token retains a spatial correspondence to an original image patch.

\noindent\textbf{Attention Weights. } \label{sec:attention_weights}
When generating a new token, the self-attention mechanism computes a weight for each preceding token. These attention weights determine the contribution of each preceding token’s value vector to the output representation of the current token at that layer. Attention weights are therefore often used as a proxy for which tokens the model is focusing on. 

\subsection{Setting}
To measure how well the model's attention aligns with the image regions referenced in its reasoning chain, we define a saliency score function. We first decompose each reasoning chain into atomic reasoning steps and retain only those that describe observations about the image.\footnote{See Appendix~\ref{app:steps_classifier} for details on reasoning-step segmentation and classification.} We compute the saliency score at the step level and apply it only to these observation steps.

\subsection{Saliency Score}
\begin{figure} [tb]
    \centering
    \includegraphics[width=0.8\textwidth]{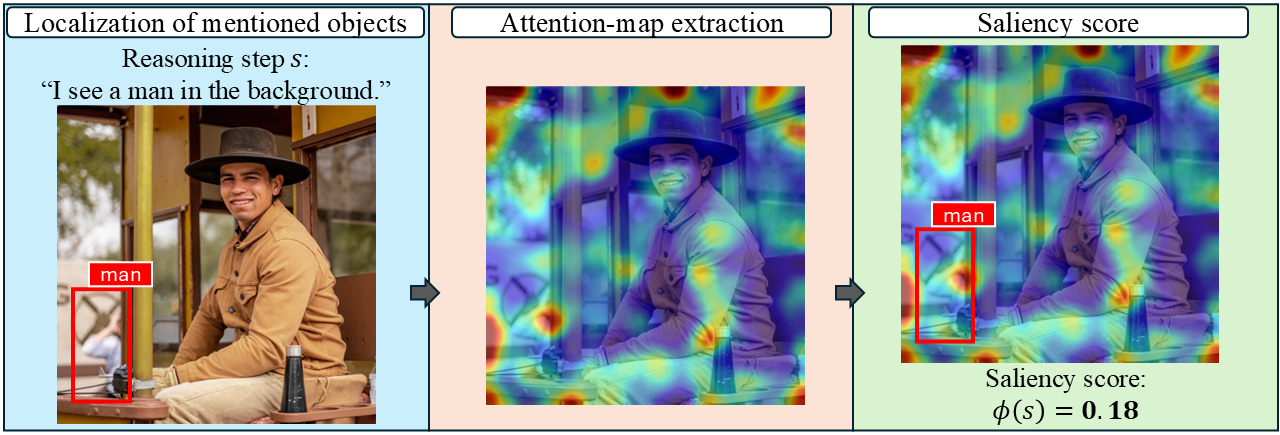}
    \caption{Our saliency-scoring pipeline. Given an input image and reasoning step, we extract bounding boxes corresponding to the mentioned image regions, compute a saliency map from the model's visual attention, and calculate the saliency score based on the overlap between the bounding boxes and the saliency map.}
    \label{fig:pipeline}
\end{figure}

Our saliency score pipeline consists of three components: localization of mentioned objects, saliency-map extraction, and saliency-map scoring. Figure~\ref{fig:pipeline} demonstrates our pipeline.

Throughout this section, we represent an image as a set of patches,
$I = \{ p_1, \ldots, p_{|I|} \}$.

\paragraph{Localization of mentioned objects.}
For each \textit{observe} step $s$, we use a phrase-grounding model on the text of the step to localize the image regions corresponding to the objects mentioned in the step. The grounding model produces a set of bounding boxes 
${b_1, \ldots, b_N}$, each represented as a set of image patches $b_i \subseteq I$.
We define the grounded image region associated with step $s$ as the union of all predicted bounding boxes: $u_s = \bigcup_{i=1}^{N} b_i$.

\paragraph{Attention-map extraction.}
Since current VLMs operate on image patches rather than individual pixels (Section~\ref{sec:patches}), we define saliency at the patch level.
We use the attention weights of selected attention heads (see Section~\ref{sec:head_selection}) as a proxy for model saliency. Specifically, for a generated token $t$ and an image patch $p$, we define $sal_t(p)$ as the attention weight assigned by $t$ to the visual token corresponding to $p$.
For a reasoning step
$s = \langle t_1, \ldots, t_{|s|} \rangle$,
we define the saliency of patch $p$ as the mean saliency across all tokens in the step:
$
sal_s(p) =
\frac{1}{|s|}
\sum_{i=1}^{|s|} sal_{t_i}(p)
$.

\paragraph{Saliency score function.}
For each observation step $s$, we combine its grounded image region $u_s$ with its patch-level saliency map $sal_s(p)$. We define the saliency score $\phi(s)$ as the mean saliency within the grounded region, normalized by the maximum saliency\footnote{We also experimented with normalizing by the mean saliency over the entire image, but found this variant to perform worse. See Appendix~\ref{app:md_baseline} for the full analysis.} over the entire image:

\begin{equation} \label{eq:phi}
\phi(s)
:=
\frac{
\frac{1}{|u_s|}
\sum_{p \in u_s} sal_s(p)
}{
\max_{p \in I} sal_s(p)
}
\end{equation}


\subsection{Rewards} \label{sec:rewards}
To apply our saliency score during training, we propose a post-training procedure based on reinforcement learning with verifiable rewards (RLVR), using GRPO~\citep{shao2024deepseekmath}. In this setting, the reward is computed deterministically from the model's generated response. The most common reward in multimodal reasoning evaluates whether the model's final answer matches the ground-truth answer~\citep{zhou2025reinforced}. We augment this objective with an additional reward that utilizes our saliency score.

We use the following reward components.
\textbf{(1) Saliency reward.}
Given a reasoning chain
$c = \langle s_1, \ldots, s_{|c|} \rangle$
and its set of extracted observation steps $O \subseteq c$, we define the saliency reward as the mean saliency score across all observation steps: $R_{sal} = \frac{1}{|O|} \sum_{s \in O} \phi(s)$.
%
%
\textbf{(2) Accuracy rewards.}
We use two complementary accuracy rewards that compare the model's final answer with the ground-truth answer: a direct-match reward, $R_{direct}$, and an LLM-as-a-judge reward, $R_{llm}$.
\textbf{(3) Format reward.} The format reward, $R_{format}$ is defined as 1 if the output is valid and 0 otherwise.
A valid output must contain a
\texttt{\textless think\textgreater}\ldots\texttt{\textless /think\textgreater}
block followed by the final answer.

The final reward is a weighted linear combination of the individual reward components mentioned above: $R
=
\alpha_{sal} R_{sal}
+
\alpha_{direct} R_{direct}
+
\alpha_{llm} R_{llm}
+
\alpha_{format} R_{format}$.


\subsection{Head Selection} \label{sec:head_selection}

Following prior work~\citep{kang2025your}, we compute the saliency reward using only a small set of selected attention heads. To identify these heads, we run the base model~\citep[Qwen3-VL-8B-Instruct;][]{bai2025qwen3} on the Visual-CoT dataset~\citep{shao2024visual} and compute the saliency reward $R_{sal}$ (Section~\ref{sec:rewards}) for each sample and attention head. We then measure, for each head, the correlation between its saliency reward and model accuracy, and select the two heads with the highest correlation.\footnote{We additionally verify that this relationship transfers across datasets by observing a similar correlation on a separate dataset.} This procedure selects heads 28 and 31 in layer 22, out of 36 layers.

This selection is consistent with the findings of \citet{gandikota2026gaze}, who show that the strongest alignment between attention and the image region being described occurs in layers 20--28. In particular, their Figure~1B shows that, for Qwen3-8B-VL, layer-wise residual-stream steering alters the model's spatial referent most strongly at layer 22.

%% file: sections/04_experiments.tex
\subsection{Training} \label{sec:training}

To bootstrap the model's reasoning capabilities
we start by training the base model, Qwen3-VL-8B-Instruct~\citep{bai2025qwen3}
using supervised fine-tuning (SFT) on the LLaVA-CoT~\citep{xu2025llava} and Mulberry-SFT~\citep{yao2026mulberry} datasets.
%
We then start GRPO training from the resulting checkpoint on the Saliency-R1-8K~\citep{gong2026saliency} dataset, holding out $100$ of its $8{,}080$ samples, using group size of $G=8$.
We set $\alpha_{format}=\alpha_{direct}=\alpha_{llm}=1$ and $\alpha_{sal}=0.4$.\footnote{Selected using a hyperparameter sweep.}

\paragraph{Auxiliary models.}
For grounding we use Grounding-DINO~\citep{liu2024grounding}, and as the LLM judge needed for $R_{llm}$ we use GPT-4o
mini~\citep{openai2024gpt4omini}.

\paragraph{Infrastructure.}
Training runs on a single node of eight
H100 GPUs in bfloat16 under DeepSpeed ZeRO-3: six GPUs hold the policy, one serves
generation with vLLM, and one hosts the grounding model. A full run takes
approximately $42$ seconds per optimizer step, or about $46$ hours.

\subsection{Evaluation} \label{sec:evaluation}

\textbf{Benchmarks.}
Recent studies have introduced a large number of visual question answering (VQA) benchmarks~\citep{kim2025visual}. However, most papers evaluate on only a small subset, raising the possibility of benchmark-selection bias. For example, VGA~\citep{zhao2026tell}, Saliency-R1~\citep{gong2026saliency}, and EASE~\citep{hu2026attend} evaluate on 3, 10, and 11 benchmarks, respectively, with only a single benchmark shared across all three. To mitigate this concern, we evaluate on all benchmarks used in recent work, together with several recently introduced benchmarks, yielding a comprehensive suite of 25 benchmarks listed in Table~\ref{tab:benchmarks}.  We evaluate with lmms-eval~\citep{zhang2025lmms}.

\input{tables/benchmarks}


\textbf{Baselines.}
We first consider several baselines that do not explicitly intervene on visual attention: the vanilla \textbf{Qwen3-VL-8B-Instruct} model, our \textbf{coldstart} checkpoint (the checkpoint after SFT we use to initialize GRPO), and a model trained using the same procedure as \name{} but without the saliency reward, i.e., using only the accuracy and format rewards (\textbf{No-Sal}). The latter tests whether the observed performance gains can be attributed simply to additional post-training rather than to the saliency objective itself.

We also compare against methods from prior work that steer visual attention toward different image regions. \textbf{VGA}~\citep{zhao2026tell} operates at inference time, steering attention toward image patches whose visual tokens are more strongly related to the objects referenced by the question. \textbf{Saliency-R1}~\citep{gong2026saliency} and \textbf{EASE}~\citep{hu2026attend} steer attention toward bounding boxes annotated by human annotators (Saliency-R1) or generated by GPT (EASE).\footnote{Since all baselines except EASE reported results on Qwen2.5, and EASE did not release trained checkpoints, we reimplemented and trained the baselines using their publicly available code. See Appendix~\ref{app:reimplementing} for further implementation details.}

\subsection{Results}
\begin{figure} [tb]
    \centering
    \includegraphics[width=0.8\textwidth]{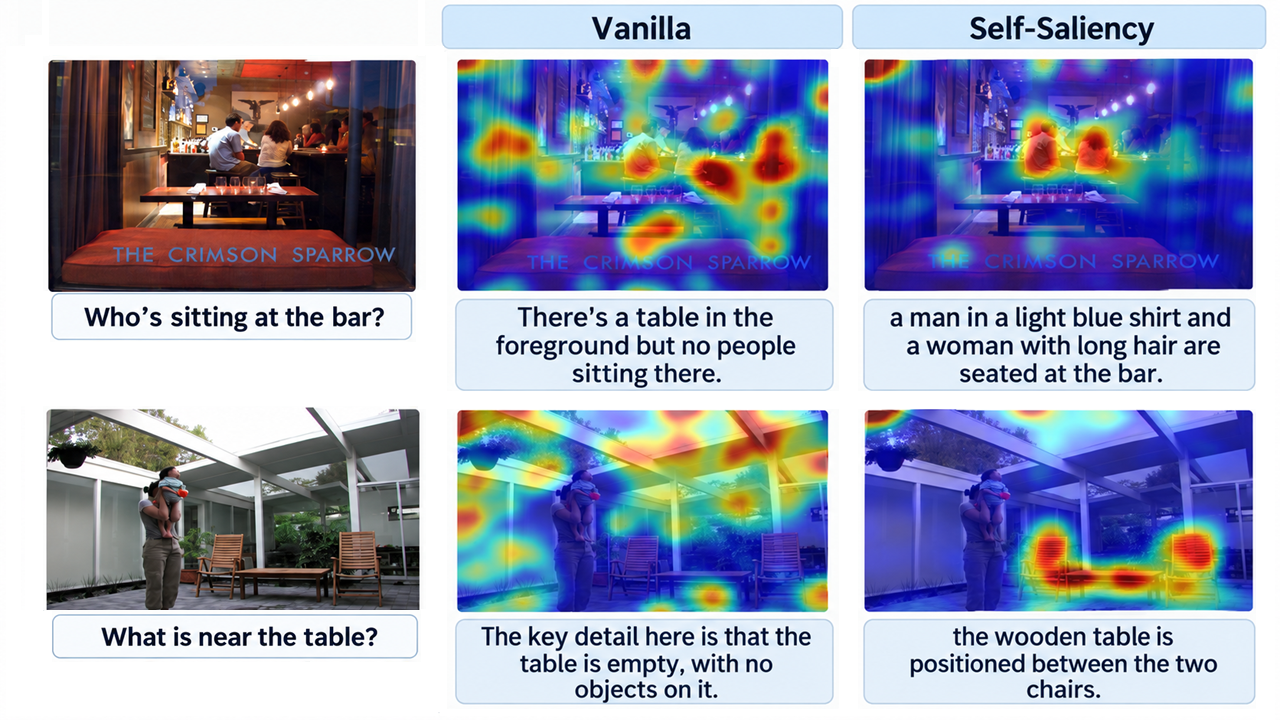}
    \caption{Qualitative examples of how \name{} correctly identifies low-level visual information.}
    \label{fig:qualitative}
\end{figure}

\input{tables/results2}
We show the results in Table~\ref{tab:main_results}. \name{} scores the highest mean score, the lowest mean rank and is most often the winner. The largest improvement is observed on benchmarks that require low-level visual saliency and multi-step reasoning (AlgoPuzzleVQA, SalBench, MME-RealWorld). Surprisingly, we also observe substantial gains on several mathematical reasoning benchmarks (WeMath, MathVision). This result is unexpected, as Grounding-DINO was trained on natural images, and we would therefore expect its localization predictions to be less reliable for mathematical diagrams. For qualitative examples of \name{}, see Figure~\ref{fig:qualitative}.

\subsection{Trained Model Analysis} \label{sec:trained_model_analysis}

\input{tables/trained_model_analysis}

The model can increase our reward in two complementary ways: by adapting its generated text to regions that are easier to ground and already highly attended, or by shifting visual attention toward the regions described in its reasoning. We investigate which mechanism dominates after training.

We compare \name{} with the coldstart checkpoint on 100 held-out validation samples. Text statistics are computed on each model's own generations. To isolate changes in attention from changes in generated text, we additionally teacher-force both models on the same set of 200 reasoning chains -- the 100 chains generated by each model -- and compute attention statistics on this fixed set.

Table~\ref{tab:trained_model_analysis} shows that \name{} generates shorter reasoning chains that refer to larger image regions. With the text held fixed, attention to grounded regions does not increase in the two heads used to compute the saliency reward. However, averaged across all heads in layer 22, grounded-region attention increases significantly ($p < 0.001$). These results suggest that \name{} primarily improves the reward by adapting its generated reasoning, while also inducing a broader shift toward grounded regions across the relevant layer.

%% file: tables/benchmarks.tex

\begin{table}[t]
\centering
\small
\setlength{\tabcolsep}{5pt}
\renewcommand{\arraystretch}{1.25}

\resizebox{0.8\columnwidth}{!}{%
\begin{tabularx}{\linewidth}{
    @{}
    >{\raggedright\arraybackslash}p{0.27\linewidth}
    >{\raggedright\arraybackslash}X
    @{}
}
\toprule
\textbf{Type} & \textbf{Benchmarks} \\
\midrule

\textbf{Mathematical reasoning} &
MathVista~\citep{lu2024mathvista},
MathVision~\citep{wang2024measuring},
MathVerse~\citep{zhang2024mathverse},
WeMath~\citep{qiao2407we},
MMK12~\citep{meng2025mm} \\

\addlinespace[2pt]
\textbf{Logical \& algorithmic puzzles} &
LogicVista~\citep{xiao2024logicvista},
AlgoPuzzleVQA~\citep{ghosal2024language},
VisuLogic~\citep{ICLR2026_2bbc73b3},
DailyClue~\citep{li2026seek} \\

\addlinespace[2pt]
\textbf{Spatial \& 3D reasoning} &
CV-Bench~\citep{zhu2025cvbench},
OmniSpatial~\citep{jia2026omnispatial} \\

\addlinespace[2pt]
\textbf{High-resolution visual search} &
V$^{*}$~\citep{wu2024v},
HR-Bench~4K\&8K~\citep{wang2025divide},
MME-RealWorld~\citep{zhang2025mme} \\

\addlinespace[2pt]
\textbf{Hallucination \& visual illusions} &
POPE~\citep{li2023evaluating},
HallusionBench~\citep{guan2024hallusionbench},
IllusionVQA~\citep{shahgir2024illusionvqa} \\

\addlinespace[2pt]
\textbf{Fine-grained perception \& chart reading} &
ChartQA~\citep{masry2022chartqa},
SalBench~\citep{huynh2025vision} \\

\addlinespace[2pt]
\textbf{General \& knowledge-intensive VQA} &
MMStar~\citep{chen2024we},
MME~\citep{fu2026mme},
RealWorldQA~\citep{xai2024grok15v},
MMMU-Pro~\citep{yue2025mmmu},
ScienceQA-IMG~\citep{lu2022learn} \\

\bottomrule
\end{tabularx}%
}

\caption{The 25 benchmarks in our evaluation suite, grouped by the capability
they primarily target.}
\label{tab:benchmarks}
\end{table}

%% file: tables/results2.tex

\newif\ifBenchRows
\BenchRowstrue

\newif\ifShowStd
\ShowStdfalse

\newif\ifDropBench
\DropBenchfalse

\providecommand{\name}{Self-Saliency}  
\newcommand{\V}[2]{#1\ifShowStd\,{\tiny$\pm$#2}\fi}      
\newcommand{\B}[2]{\textbf{#1}\ifShowStd\,{\tiny$\pm$#2}\fi} 
\newcommand{\Vn}[1]{#1}                                   
\newcommand{\Bn}[1]{\textbf{#1}}                          
\newcommand{\up}{$\uparrow$}
\newcommand{\down}{$\downarrow$}
\newcommand{\NBench}{\ifDropBench 21\else 25\fi}

\ifBenchRows
\begin{table*}[t]
\centering
\caption{Results across \NBench{} benchmarks. Best per benchmark in \textbf{bold}. Mean score is computed over all \NBench{} benchmarks after rescaling MME to $[0,100]$ (raw score $/\,2800$). Mean rank is the average over per-benchmark ranks (1 = best).\ifDropBench\ DailyClue, HallusionBench, MMK12 and VisuLogic are excluded.\fi  }
\label{tab:main_results}
\resizebox{\textwidth}{!}{%
\begin{tabular}{l ccc ccc c}
\toprule
& \multicolumn{3}{c}{\textit{No attention intervention}} & \multicolumn{3}{c}{\textit{Prior work}} & \multicolumn{1}{c}{\textit{Ours}} \\
\cmidrule(lr){2-4} \cmidrule(lr){5-7} \cmidrule(lr){8-8}
Benchmark & \makecell{Qwen3-VL\\8B-Instruct} & Coldstart & No-Sal & VGA & \makecell{Saliency\\R1} & EASE & \name{} \\
\midrule
AlgoPuzzleVQA \up & \V{32.8}{1.11} & \V{36.9}{1.14} & \V{39.9}{1.15} & \V{34.3}{1.12} & \V{38.7}{1.15} & \V{36.4}{1.13} & \B{41.4}{1.16} \\
ChartQA \up & \V{85.0}{0.71} & \V{88.8}{0.63} & \V{89.4}{0.62} & \V{84.9}{0.72} & \V{89.7}{0.61} & \V{88.9}{0.63} & \B{90.0}{0.60} \\
CV-Bench \up & \V{86.0}{0.68} & \V{83.4}{0.72} & \V{84.9}{0.70} & \B{86.1}{0.67} & \V{83.8}{0.72} & \V{84.2}{0.71} & \V{85.9}{0.68} \\
\ifDropBench\else DailyClue \up & \B{35.4}{1.85} & \V{32.4}{1.82} & \V{34.2}{1.84} & \V{34.2}{1.84} & \V{32.0}{1.81} & \V{32.7}{1.82} & \V{31.2}{1.80} \\ \fi
\ifDropBench\else HallusionBench \up & \Bn{63.2} & \Vn{59.7} & \Vn{57.1} & \Vn{62.0} & \Vn{58.3} & \Vn{62.6} & \Vn{55.7} \\ \fi
HRBench4k \up & \V{78.0}{2.70} & \V{76.9}{2.70} & \V{77.4}{2.63} & \V{78.1}{2.68} & \V{78.9}{2.57} & \V{76.9}{2.69} & \B{79.5}{2.60} \\
HRBench8k \up & \V{74.1}{2.75} & \V{72.9}{2.77} & \V{72.0}{2.79} & \V{74.2}{2.77} & \B{76.0}{2.66} & \V{72.8}{2.86} & \V{74.5}{2.73} \\
IllusionVQA \up & \V{40.0}{1.55} & \V{41.9}{1.56} & \V{41.5}{1.56} & \V{40.1}{1.55} & \B{43.8}{1.57} & \V{40.5}{1.55} & \V{43.5}{1.57} \\
LogicVista \up & \V{52.5}{2.36} & \V{54.9}{2.35} & \V{55.1}{2.35} & \V{51.3}{2.36} & \B{56.5}{2.35} & \V{52.9}{2.36} & \V{56.2}{2.35} \\
MathVerse \up & \V{34.4}{1.69} & \V{51.0}{1.78} & \B{52.8}{1.78} & \V{33.8}{1.69} & \V{49.5}{1.78} & \V{50.4}{1.78} & \V{50.9}{1.78} \\
MathVision \up & \V{29.6}{2.62} & \V{33.5}{2.71} & \V{37.5}{2.78} & \V{31.2}{2.66} & \V{33.9}{2.72} & \V{35.9}{2.76} & \B{41.5}{2.83} \\
MathVista \up & \V{73.5}{1.40} & \V{69.5}{1.46} & \V{71.8}{1.42} & \B{73.8}{1.39} & \V{71.0}{1.44} & \V{70.0}{1.45} & \V{70.4}{1.44} \\
MME \up & \V{2375.9}{36.2} & \V{2371.7}{35.1} & \V{2376.1}{36.3} & \V{2379.8}{36.2} & \B{2395.7}{34.0} & \V{2395.4}{33.1} & \V{2355.1}{35.1} \\
MME-RW \up & \V{63.7}{0.31} & \V{64.2}{0.31} & \V{66.7}{0.31} & \V{63.8}{0.31} & \V{66.4}{0.31} & \V{65.1}{0.31} & \B{68.7}{0.30} \\
\ifDropBench\else MMK12 \up & \B{68.2}{1.04} & \V{63.1}{1.08} & \V{62.3}{1.08} & \V{67.9}{1.04} & \V{61.1}{1.09} & \V{63.1}{1.08} & \V{62.0}{1.09} \\ \fi
MMMUPro \up & \V{40.4}{1.18} & \V{48.6}{1.20} & \B{48.6}{1.20} & \V{41.9}{1.19} & \V{47.7}{1.20} & \V{47.1}{1.20} & \V{48.4}{1.20} \\
MMStar \up & \V{64.0}{1.34} & \V{69.8}{1.29} & \B{70.9}{1.26} & \V{64.7}{1.33} & \V{69.9}{1.28} & \V{69.5}{1.28} & \V{69.6}{1.28} \\
Omnispatial \up & \V{47.4}{1.28} & \V{50.3}{1.28} & \V{48.2}{1.28} & \V{46.8}{1.27} & \V{50.2}{1.28} & \V{50.2}{1.28} & \B{51.0}{1.28} \\
SalBench \up & \V{43.3}{0.97} & \V{60.7}{0.96} & \V{59.8}{0.96} & \V{43.3}{0.97} & \V{62.7}{0.95} & \V{58.6}{0.97} & \B{64.7}{0.94} \\
POPE \up & \V{88.8}{0.33} & \V{87.4}{0.35} & \V{88.2}{0.34} & \V{88.8}{0.33} & \V{88.5}{0.34} & \V{88.5}{0.34} & \B{88.9}{0.33} \\
RealWorldQA \up & \V{68.5}{1.68} & \V{67.1}{1.70} & \V{67.8}{1.69} & \V{68.4}{1.68} & \B{68.8}{1.68} & \V{67.5}{1.70} & \V{68.2}{1.68} \\
ScienceQA \up & \V{94.4}{0.51} & \V{96.3}{0.42} & \V{97.2}{0.37} & \V{94.4}{0.51} & \B{97.6}{0.34} & \V{96.2}{0.42} & \V{97.4}{0.35} \\
\ifDropBench\else VisuLogic \up & \B{26.6}{1.40} & \V{24.9}{1.37} & \V{24.8}{1.37} & \V{26.4}{1.39} & \V{24.9}{1.37} & \V{24.4}{1.36} & \V{24.4}{1.36} \\ \fi
V$^{*}$ \up & \V{84.3}{2.64} & \V{79.6}{2.92} & \V{79.6}{2.92} & \B{85.3}{2.57} & \V{80.6}{2.87} & \V{81.2}{2.84} & \V{82.7}{2.74} \\
WeMath \up & \V{69.4}{1.10} & \V{68.6}{1.11} & \V{74.3}{1.05} & \V{68.2}{1.12} & \V{73.8}{1.05} & \V{71.1}{1.09} & \B{75.5}{1.03} \\
\midrule
Mean score \up & \ifDropBench \Vn{63.6}\else \Vn{61.1}\fi & \ifDropBench \Vn{66.0}\else \Vn{62.7}\fi & \ifDropBench \Vn{67.1}\else \Vn{63.5}\fi & \ifDropBench \Vn{63.7}\else \Vn{61.2}\fi & \ifDropBench \Vn{67.3}\else \Vn{63.6}\fi & \ifDropBench \Vn{66.2}\else \Vn{62.9}\fi & \ifDropBench \Bn{68.2}\else \Bn{64.3}\fi \\
Mean rank \down & \ifDropBench \Vn{5.07}\else \Vn{4.42}\fi & \ifDropBench \Vn{4.76}\else \Vn{4.64}\fi & \ifDropBench \Vn{3.50}\else \Vn{3.68}\fi & \ifDropBench \Vn{4.88}\else \Vn{4.48}\fi & \ifDropBench \Vn{2.86}\else \Vn{3.26}\fi & \ifDropBench \Vn{4.60}\else \Vn{4.50}\fi & \ifDropBench \Bn{2.33}\else \Bn{3.02}\fi \\
Number of Wins \up & \ifDropBench \Vn{0}\else \Vn{4}\fi & \Vn{0} & \Vn{3} & \Vn{3} & \Vn{6} & \Vn{0} & \Bn{9} \\
\bottomrule
\end{tabular}}
\end{table*}

\else
\ifDropBench
\begin{table*}[t]
\centering
\caption{Results across \NBench{} benchmarks (part 1 of 3). Best per benchmark in \textbf{bold}.}
\label{tab:main_results}
\resizebox{\textwidth}{!}{%
\begin{tabular}{lccccccc}
\toprule
Method & AlgoPuzzleVQA \up & ChartQA \up & CV-Bench \up & HRBench \up & HRBench \up & IllusionVQA \up & LogicVista \up \\
 &  &  &  & {\tiny\makecell{4k}} & {\tiny\makecell{8k}} & {\tiny\makecell{soft\\localization}} &  \\
\midrule
\multicolumn{8}{l}{\textit{No attention intervention}} \\
\quad Qwen3-VL-8B-Instruct & \V{32.8}{1.11} & \V{85.0}{0.71} & \V{86.0}{0.68} & \V{78.0}{2.70} & \V{74.1}{2.75} & \V{40.0}{1.55} & \V{52.5}{2.36} \\
\quad Coldstart & \V{36.9}{1.14} & \V{88.8}{0.63} & \V{83.4}{0.72} & \V{76.9}{2.70} & \V{72.9}{2.77} & \V{41.9}{1.56} & \V{54.9}{2.35} \\
\quad No-Sal & \V{39.9}{1.15} & \V{89.4}{0.62} & \V{84.9}{0.70} & \V{77.4}{2.63} & \V{72.0}{2.79} & \V{41.5}{1.56} & \V{55.1}{2.35} \\
\midrule
\multicolumn{8}{l}{\textit{Prior work}} \\
\quad VGA & \V{34.3}{1.12} & \V{84.9}{0.72} & \B{86.1}{0.67} & \V{78.1}{2.68} & \V{74.2}{2.77} & \V{40.1}{1.55} & \V{51.3}{2.36} \\
\quad Saliency-R1 & \V{38.7}{1.15} & \V{89.7}{0.61} & \V{83.8}{0.72} & \V{78.9}{2.57} & \B{76.0}{2.66} & \B{43.8}{1.57} & \B{56.5}{2.35} \\
\quad EASE & \V{36.4}{1.13} & \V{88.9}{0.63} & \V{84.2}{0.71} & \V{76.9}{2.69} & \V{72.8}{2.86} & \V{40.5}{1.55} & \V{52.9}{2.36} \\
\midrule
\multicolumn{8}{l}{\textit{Ours}} \\
\quad \name{} & \B{41.4}{1.16} & \B{90.0}{0.60} & \V{85.9}{0.68} & \B{79.5}{2.60} & \V{74.5}{2.73} & \V{43.5}{1.57} & \V{56.2}{2.35} \\
\bottomrule
\end{tabular}}
\end{table*}

\begin{table*}[t]
\centering
\caption{Results across \NBench{} benchmarks (part 2 of 3). Best per benchmark in \textbf{bold}.}
\label{tab:main_results}
\resizebox{\textwidth}{!}{%
\begin{tabular}{lccccccc}
\toprule
Method & MathVerse \up & MathVision \up & MathVista \up & MME \up & MME-RW \up & MMMUPro \up & MMStar \up \\
 & {\tiny\makecell{testmini\\vision\_only}} & {\tiny\makecell{testmini}} & {\tiny\makecell{testmini}} &  &  & {\tiny\makecell{standard}} &  \\
\midrule
\multicolumn{8}{l}{\textit{No attention intervention}} \\
\quad Qwen3-VL-8B-Instruct & \V{34.4}{1.69} & \V{29.6}{2.62} & \V{73.5}{1.40} & \V{2375.9}{36.2} & \V{63.7}{0.31} & \V{40.4}{1.18} & \V{64.0}{1.34} \\
\quad Coldstart & \V{51.0}{1.78} & \V{33.5}{2.71} & \V{69.5}{1.46} & \V{2371.7}{35.1} & \V{64.2}{0.31} & \V{48.6}{1.20} & \V{69.8}{1.29} \\
\quad No-Sal & \B{52.8}{1.78} & \V{37.5}{2.78} & \V{71.8}{1.42} & \V{2376.1}{36.3} & \V{66.7}{0.31} & \B{48.6}{1.20} & \B{70.9}{1.26} \\
\midrule
\multicolumn{8}{l}{\textit{Prior work}} \\
\quad VGA & \V{33.8}{1.69} & \V{31.2}{2.66} & \B{73.8}{1.39} & \V{2379.8}{36.2} & \V{63.8}{0.31} & \V{41.9}{1.19} & \V{64.7}{1.33} \\
\quad Saliency-R1 & \V{49.5}{1.78} & \V{33.9}{2.72} & \V{71.0}{1.44} & \B{2395.7}{34.0} & \V{66.4}{0.31} & \V{47.7}{1.20} & \V{69.9}{1.28} \\
\quad EASE & \V{50.4}{1.78} & \V{35.9}{2.76} & \V{70.0}{1.45} & \V{2395.4}{33.1} & \V{65.1}{0.31} & \V{47.1}{1.20} & \V{69.5}{1.28} \\
\midrule
\multicolumn{8}{l}{\textit{Ours}} \\
\quad \name{} & \V{50.9}{1.78} & \B{41.5}{2.83} & \V{70.4}{1.44} & \V{2355.1}{35.1} & \B{68.7}{0.30} & \V{48.4}{1.20} & \V{69.6}{1.28} \\
\bottomrule
\end{tabular}}
\end{table*}

\begin{table*}[t]
\centering
\caption{Results across \NBench{} benchmarks (part 3 of 3), with the aggregate columns. Best per benchmark in \textbf{bold}. Mean score is computed over all \NBench{} benchmarks after rescaling MME to $[0,100]$ (raw score $/\,2800$); mean rank averages the per-benchmark ranks (1 = best).\ifDropBench\ DailyClue, HallusionBench, MMK12 and VisuLogic are excluded.\fi}
\label{tab:main_results}
\resizebox{\textwidth}{!}{%
\begin{tabular}{lccccccc|cc}
\toprule
Method & Omnispatial \up & SalBench \up & POPE \up & RealWorldQA \up & ScienceQA \up & V$^{*}$ \up & WeMath \up & \makecell{Mean\\score} \up & \makecell{Mean\\rank} \down \\
 & {\tiny\makecell{test}} & {\tiny\makecell{p3}} &  &  & {\tiny\makecell{img}} &  & {\tiny\makecell{testmini}} & &  \\
\midrule
\multicolumn{10}{l}{\textit{No attention intervention}} \\
\quad Qwen3-VL-8B-Instruct & \V{47.4}{1.28} & \V{43.3}{0.97} & \V{88.8}{0.33} & \V{68.5}{1.68} & \V{94.4}{0.51} & \V{84.3}{2.64} & \V{69.4}{1.10} & \ifDropBench \Vn{63.6}\else \Vn{61.1}\fi & \ifDropBench \Vn{5.07}\else \Vn{4.42}\fi \\
\quad Coldstart & \V{50.3}{1.28} & \V{60.7}{0.96} & \V{87.4}{0.35} & \V{67.1}{1.70} & \V{96.3}{0.42} & \V{79.6}{2.92} & \V{68.6}{1.11} & \ifDropBench \Vn{66.0}\else \Vn{62.7}\fi & \ifDropBench \Vn{4.76}\else \Vn{4.64}\fi \\
\quad No-Sal & \V{48.2}{1.28} & \V{59.8}{0.96} & \V{88.2}{0.34} & \V{67.8}{1.69} & \V{97.2}{0.37} & \V{79.6}{2.92} & \V{74.3}{1.05} & \ifDropBench \Vn{67.1}\else \Vn{63.5}\fi & \ifDropBench \Vn{3.50}\else \Vn{3.68}\fi \\
\midrule
\multicolumn{10}{l}{\textit{Prior work}} \\
\quad VGA & \V{46.8}{1.27} & \V{43.3}{0.97} & \V{88.8}{0.33} & \V{68.4}{1.68} & \V{94.4}{0.51} & \B{85.3}{2.57} & \V{68.2}{1.12} & \ifDropBench \Vn{63.7}\else \Vn{61.2}\fi & \ifDropBench \Vn{4.88}\else \Vn{4.48}\fi \\
\quad Saliency-R1 & \V{50.2}{1.28} & \V{62.7}{0.95} & \V{88.5}{0.34} & \B{68.8}{1.68} & \B{97.6}{0.34} & \V{80.6}{2.87} & \V{73.8}{1.05} & \ifDropBench \Vn{67.3}\else \Vn{63.6}\fi & \ifDropBench \Vn{2.86}\else \Vn{3.26}\fi \\
\quad EASE & \V{50.2}{1.28} & \V{58.6}{0.97} & \V{88.5}{0.34} & \V{67.5}{1.70} & \V{96.2}{0.42} & \V{81.2}{2.84} & \V{71.1}{1.09} & \ifDropBench \Vn{66.2}\else \Vn{62.9}\fi & \ifDropBench \Vn{4.60}\else \Vn{4.50}\fi \\
\midrule
\multicolumn{10}{l}{\textit{Ours}} \\
\quad \name{} & \B{51.0}{1.28} & \B{64.7}{0.94} & \B{88.9}{0.33} & \V{68.2}{1.68} & \V{97.4}{0.35} & \V{82.7}{2.74} & \B{75.5}{1.03} & \ifDropBench \Bn{68.2}\else \Bn{64.3}\fi & \ifDropBench \Bn{2.33}\else \Bn{3.02}\fi \\
\bottomrule
\end{tabular}}
\end{table*}

\else
\begin{table*}[t]
\centering
\caption{Results across \NBench{} benchmarks (part 1 of 3). Best per benchmark in \textbf{bold}.}
\label{tab:main_results}
\resizebox{\textwidth}{!}{%
\begin{tabular}{lccccccccc}
\toprule
Method & AlgoPuzzleVQA \up & ChartQA \up & CV-Bench \up & DailyClue \up & HallusionBench \up & HRBench \up & HRBench \up & IllusionVQA \up & LogicVista \up \\
 &  &  &  &  & {\tiny\makecell{image}} & {\tiny\makecell{4k}} & {\tiny\makecell{8k}} & {\tiny\makecell{soft\\localization}} &  \\
\midrule
\multicolumn{10}{l}{\textit{No attention intervention}} \\
\quad Qwen3-VL-8B-Instruct & \V{32.8}{1.11} & \V{85.0}{0.71} & \V{86.0}{0.68} & \B{35.4}{1.85} & \Bn{63.2} & \V{78.0}{2.70} & \V{74.1}{2.75} & \V{40.0}{1.55} & \V{52.5}{2.36} \\
\quad Coldstart & \V{36.9}{1.14} & \V{88.8}{0.63} & \V{83.4}{0.72} & \V{32.4}{1.82} & \Vn{59.7} & \V{76.9}{2.70} & \V{72.9}{2.77} & \V{41.9}{1.56} & \V{54.9}{2.35} \\
\quad No-Sal & \V{39.9}{1.15} & \V{89.4}{0.62} & \V{84.9}{0.70} & \V{34.2}{1.84} & \Vn{57.1} & \V{77.4}{2.63} & \V{72.0}{2.79} & \V{41.5}{1.56} & \V{55.1}{2.35} \\
\midrule
\multicolumn{10}{l}{\textit{Prior work}} \\
\quad VGA & \V{34.3}{1.12} & \V{84.9}{0.72} & \B{86.1}{0.67} & \V{34.2}{1.84} & \Vn{62.0} & \V{78.1}{2.68} & \V{74.2}{2.77} & \V{40.1}{1.55} & \V{51.3}{2.36} \\
\quad Saliency-R1 & \V{38.7}{1.15} & \V{89.7}{0.61} & \V{83.8}{0.72} & \V{32.0}{1.81} & \Vn{58.3} & \V{78.9}{2.57} & \B{76.0}{2.66} & \B{43.8}{1.57} & \B{56.5}{2.35} \\
\quad EASE & \V{36.4}{1.13} & \V{88.9}{0.63} & \V{84.2}{0.71} & \V{32.7}{1.82} & \Vn{62.6} & \V{76.9}{2.69} & \V{72.8}{2.86} & \V{40.5}{1.55} & \V{52.9}{2.36} \\
\midrule
\multicolumn{10}{l}{\textit{Ours}} \\
\quad \name{} & \B{41.4}{1.16} & \B{90.0}{0.60} & \V{85.9}{0.68} & \V{31.2}{1.80} & \Vn{55.7} & \B{79.5}{2.60} & \V{74.5}{2.73} & \V{43.5}{1.57} & \V{56.2}{2.35} \\
\bottomrule
\end{tabular}}
\end{table*}

\begin{table*}[t]
\centering
\caption{Results across \NBench{} benchmarks (part 2 of 3). Best per benchmark in \textbf{bold}.}
\label{tab:main_results}
\resizebox{\textwidth}{!}{%
\begin{tabular}{lcccccccc}
\toprule
Method & MathVerse \up & MathVision \up & MathVista \up & MME \up & MME-RW \up & MMK12 \up & MMMUPro \up & MMStar \up \\
 & {\tiny\makecell{testmini\\vision\_only}} & {\tiny\makecell{testmini}} & {\tiny\makecell{testmini}} &  &  &  & {\tiny\makecell{standard}} &  \\
\midrule
\multicolumn{9}{l}{\textit{No attention intervention}} \\
\quad Qwen3-VL-8B-Instruct & \V{34.4}{1.69} & \V{29.6}{2.62} & \V{73.5}{1.40} & \V{2375.9}{36.2} & \V{63.7}{0.31} & \B{68.2}{1.04} & \V{40.4}{1.18} & \V{64.0}{1.34} \\
\quad Coldstart & \V{51.0}{1.78} & \V{33.5}{2.71} & \V{69.5}{1.46} & \V{2371.7}{35.1} & \V{64.2}{0.31} & \V{63.1}{1.08} & \V{48.6}{1.20} & \V{69.8}{1.29} \\
\quad No-Sal & \B{52.8}{1.78} & \V{37.5}{2.78} & \V{71.8}{1.42} & \V{2376.1}{36.3} & \V{66.7}{0.31} & \V{62.3}{1.08} & \B{48.6}{1.20} & \B{70.9}{1.26} \\
\midrule
\multicolumn{9}{l}{\textit{Prior work}} \\
\quad VGA & \V{33.8}{1.69} & \V{31.2}{2.66} & \B{73.8}{1.39} & \V{2379.8}{36.2} & \V{63.8}{0.31} & \V{67.9}{1.04} & \V{41.9}{1.19} & \V{64.7}{1.33} \\
\quad Saliency-R1 & \V{49.5}{1.78} & \V{33.9}{2.72} & \V{71.0}{1.44} & \B{2395.7}{34.0} & \V{66.4}{0.31} & \V{61.1}{1.09} & \V{47.7}{1.20} & \V{69.9}{1.28} \\
\quad EASE & \V{50.4}{1.78} & \V{35.9}{2.76} & \V{70.0}{1.45} & \V{2395.4}{33.1} & \V{65.1}{0.31} & \V{63.1}{1.08} & \V{47.1}{1.20} & \V{69.5}{1.28} \\
\midrule
\multicolumn{9}{l}{\textit{Ours}} \\
\quad \name{} & \V{50.9}{1.78} & \B{41.5}{2.83} & \V{70.4}{1.44} & \V{2355.1}{35.1} & \B{68.7}{0.30} & \V{62.0}{1.09} & \V{48.4}{1.20} & \V{69.6}{1.28} \\
\bottomrule
\end{tabular}}
\end{table*}

\begin{table*}[t]
\centering
\caption{Results across \NBench{} benchmarks (part 3 of 3), with the aggregate columns. Best per benchmark in \textbf{bold}. Mean score is computed over all \NBench{} benchmarks after rescaling MME to $[0,100]$ (raw score $/\,2800$); mean rank averages the per-benchmark ranks (1 = best).\ifDropBench\ DailyClue, HallusionBench, MMK12 and VisuLogic are excluded.\fi}
\label{tab:main_results}
\resizebox{\textwidth}{!}{%
\begin{tabular}{lcccccccc|cc}
\toprule
Method & Omnispatial \up & SalBench \up & POPE \up & RealWorldQA \up & ScienceQA \up & VisuLogic \up & V$^{*}$ \up & WeMath \up & \makecell{Mean\\score} \up & \makecell{Mean\\rank} \down \\
 & {\tiny\makecell{test}} & {\tiny\makecell{p3}} &  &  & {\tiny\makecell{img}} &  &  & {\tiny\makecell{testmini}} & &  \\
\midrule
\multicolumn{11}{l}{\textit{No attention intervention}} \\
\quad Qwen3-VL-8B-Instruct & \V{47.4}{1.28} & \V{43.3}{0.97} & \V{88.8}{0.33} & \V{68.5}{1.68} & \V{94.4}{0.51} & \B{26.6}{1.40} & \V{84.3}{2.64} & \V{69.4}{1.10} & \ifDropBench \Vn{63.6}\else \Vn{61.1}\fi & \ifDropBench \Vn{5.07}\else \Vn{4.42}\fi \\
\quad Coldstart & \V{50.3}{1.28} & \V{60.7}{0.96} & \V{87.4}{0.35} & \V{67.1}{1.70} & \V{96.3}{0.42} & \V{24.9}{1.37} & \V{79.6}{2.92} & \V{68.6}{1.11} & \ifDropBench \Vn{66.0}\else \Vn{62.7}\fi & \ifDropBench \Vn{4.76}\else \Vn{4.64}\fi \\
\quad No-Sal & \V{48.2}{1.28} & \V{59.8}{0.96} & \V{88.2}{0.34} & \V{67.8}{1.69} & \V{97.2}{0.37} & \V{24.8}{1.37} & \V{79.6}{2.92} & \V{74.3}{1.05} & \ifDropBench \Vn{67.1}\else \Vn{63.5}\fi & \ifDropBench \Vn{3.50}\else \Vn{3.68}\fi \\
\midrule
\multicolumn{11}{l}{\textit{Prior work}} \\
\quad VGA & \V{46.8}{1.27} & \V{43.3}{0.97} & \V{88.8}{0.33} & \V{68.4}{1.68} & \V{94.4}{0.51} & \V{26.4}{1.39} & \B{85.3}{2.57} & \V{68.2}{1.12} & \ifDropBench \Vn{63.7}\else \Vn{61.2}\fi & \ifDropBench \Vn{4.88}\else \Vn{4.48}\fi \\
\quad Saliency-R1 & \V{50.2}{1.28} & \V{62.7}{0.95} & \V{88.5}{0.34} & \B{68.8}{1.68} & \B{97.6}{0.34} & \V{24.9}{1.37} & \V{80.6}{2.87} & \V{73.8}{1.05} & \ifDropBench \Vn{67.3}\else \Vn{63.6}\fi & \ifDropBench \Vn{2.86}\else \Vn{3.26}\fi \\
\quad EASE & \V{50.2}{1.28} & \V{58.6}{0.97} & \V{88.5}{0.34} & \V{67.5}{1.70} & \V{96.2}{0.42} & \V{24.4}{1.36} & \V{81.2}{2.84} & \V{71.1}{1.09} & \ifDropBench \Vn{66.2}\else \Vn{62.9}\fi & \ifDropBench \Vn{4.60}\else \Vn{4.50}\fi \\
\midrule
\multicolumn{11}{l}{\textit{Ours}} \\
\quad \name{} & \B{51.0}{1.28} & \B{64.7}{0.94} & \B{88.9}{0.33} & \V{68.2}{1.68} & \V{97.4}{0.35} & \V{24.4}{1.36} & \V{82.7}{2.74} & \B{75.5}{1.03} & \ifDropBench \Bn{68.2}\else \Bn{64.3}\fi & \ifDropBench \Bn{2.33}\else \Bn{3.02}\fi \\
\bottomrule
\end{tabular}}
\end{table*}

\fi

\fi

%% file: tables/trained_model_analysis.tex
\begin{table}[t]
    \centering
    \caption{Text and attention statistics before and after training, averaged over 100 held-out validation samples. In-region share: attention in grounded regions divided by all visual attention. \name{} primarily adapts its generated text to the selected heads' attention patterns, while also increasing attention to grounded regions across layer 22.}
    \label{tab:trained_model_analysis}
    \setlength{\tabcolsep}{5pt}
    \renewcommand{\arraystretch}{1.15}

    \resizebox{0.7\columnwidth}{!}{%
    \begin{tabular}{l|cc|cc}
        \toprule
        & \multicolumn{2}{c|}{Text-related} & \multicolumn{2}{c}{Attention-related}
        \\
        \midrule
        & \makecell{Completion}
        & \makecell{Box union}
        & \makecell{In-region share}
        & \makecell{In-region share} \\
        & \makecell{length}
        & \makecell{area}
        & \makecell{(selected heads)}
        & \makecell{(layer-22 heads)} \\
        \midrule
        Coldstart &
        230.0 & 0.533 &  0.462 & 0.612 \\
        \name{}
        & 149.6
        & 0.566
        & 0.462
        & 0.621 \\
        \bottomrule
    \end{tabular}%
    }
\end{table}

%% file: sections/05_attention_analysis.tex
In this section, we analyze the visual attention patterns of commonly used VLMs to better understand the training signal provided by \name{}. 
Prior work has suggested that visual attention tends to concentrate on background regions~\citep{gong2024damro, darcet2024vision} or regions that are visually confusing or semantically related to the target object~\citep{wang2026same}.\footnote{LLaVA was shown to attend to the lower portion of the image~\citep{zhang2025beyond,zhao2025mca}; here we focus on other models.} We propose a framing that does not contradict previous findings, but is more useful for our purposes: VLM attention is disproportionately concentrated along the geometric border of the image.

\subsection{Experimental Setup} \label{sec:attention_exp}

We use 1,800 (image, question) pairs from the Visual-CoT dataset~\citep{shao2024visual} to prompt four released instruction-tuned VLMs: Qwen3-VL-8B-Instruct~\citep{bai2025qwen3}, InternVL3.5-8B~\citep{wang2025internvl3}, GLM-4.1V-Thinking~\citep{hong2025glm} and Nemotron-Omni~\citep{deshmukh2026nemotron}.\footnote{See Appendix~\ref{app:model_details} for a list of each model's characteristics.} For every image we run a forward pass and record, at every layer and head, the attention each image patch receives, averaged over the generated tokens
-- and renormalized within the image.\footnote{For models with image-dependent grid shape, we take the most common grid shape, and use area-weighted averaging to map all grids to a fixed-size averaged grid.}

Given a set of patches, we define its \emph{enrichment} ($En$) as the fraction of total visual attention assigned to the set, normalized by the fraction of image patches it contains. An enrichment of $En=1$ indicates no effect, while $En>1$ and $En<1$ indicate disproportionately high and low attention, respectively.
We record two sets of patches: the border ring (all patches on the border of the image) and the center (all patches but the border ring).






\input{tables/enrichment}
\subsection{Results} \label{sec:attention_exp_results}
\begin{figure} [tb]
    \centering
    \includegraphics[width=0.8\textwidth]{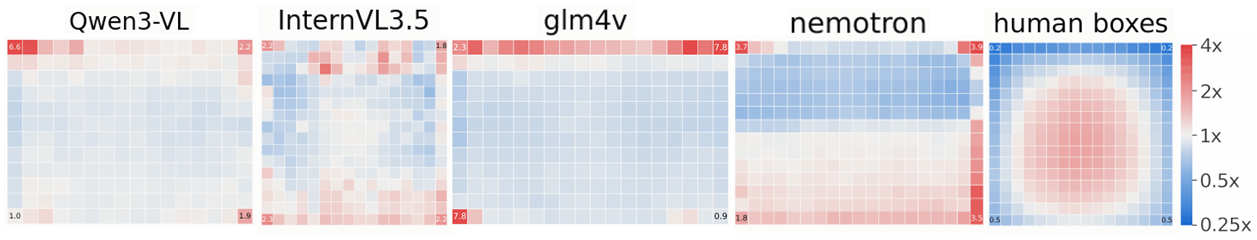}
    \caption{Saliency maps for different models and human-annotated bounding boxes. For each model, each cell represents the fraction of the model's total visual attention assigned to the corresponding image patch. For the human annotations (right), each cell represents the fraction of images in which the corresponding patch falls within a human-annotated bounding box.}
    \label{fig:enrichment}
\end{figure}
Table~\ref{tab:enrichment} shows enrichment results. 
%
Across all models, attention is more enriched along the border ring than in the center of the image. Moreover, in every model, the patch with the highest enrichment is located at a corner of the grid.
Figure~\ref{fig:enrichment} visualizes the enrichment of each image patch for every model.

\paragraph{Relevant image areas are in the center.}
We use the bounding-box annotations from Visual-CoT, in which human annotators identify the image regions they consider important for answering each question. Over the same set of images, we compute an analogous patch-level statistic: we measure the fraction of images in which each patch falls within an annotated bounding box. The bottom row of Table~\ref{tab:enrichment} and the rightmost plot in Figure~\ref{fig:enrichment} present the results. While models allocate more attention to image boundaries, human annotations indicate that regions near the center are more relevant for answering questions.

\paragraph{The most attended patch is typically outside the grounded area.} Because attention peaks near image corners while relevant regions tend to lie near the center, the maximum-attention patch usually falls outside the grounded region. Consequently, increasing attention within the grounded area raises the numerator of $\phi$ (Equation~\ref{eq:phi}) without increasing its max-normalized denominator.




\begin{figure} [tb]
    \centering
    \includegraphics[width=0.4\textwidth]{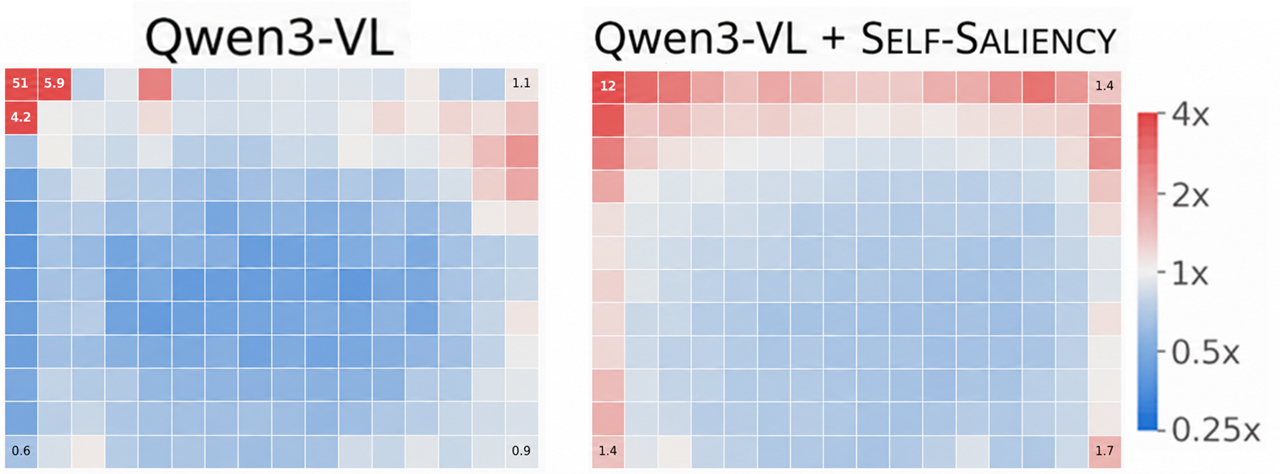}
    \caption{The fraction of the model's total visual attention assigned to the corresponding image patch, for the heads used to extract saliency (layer 22, heads 28+31), before and after training.}
    \label{fig:before_and_after}
\end{figure}

\paragraph{\name{} reduces corner-focused attention.}
Figure~\ref{fig:before_and_after} shows the average attention heatmaps of the two selected heads before and after training. After training, attention is redistributed primarily from corner sinks to the image border, with a smaller fraction moving toward the center.

\subsection{Ablation Study}

To isolate the contribution of different components of our method, we train two variants using the same procedure as \name{}, while changing only the source of the target bounding boxes. Motivated by the findings in Section~\ref{sec:attention_exp_results}, \textbf{center\_rect} uses a fixed rectangle centered in the image. If this variant performs comparably to \name{}, it would suggest that the gains primarily result from shifting attention away from the grid corner. In contrast, \textbf{question\_boxes} applies the grounding model to the input question rather than to the model's reasoning steps, allowing us to test whether chain-level grounding provides additional benefit over grounding image-level text.



\input{tables/ablation_results2}

We present the results in Table~\ref{tab:ablation_results} and show a clear advantage for \name{} over the ablation variants. This finding indicates that grounding the model's reasoning chain provides a meaningful signal for alignment with visual attention.



%% file: tables/enrichment.tex
\begin{table}[t]
    \centering
    \caption{Attention enrichment for different models.}
    \label{tab:enrichment}

    \resizebox{0.7\columnwidth}{!}{%
    \begin{tabular}{lccc}
        \toprule
        Model & $En$ border ring & $En$ center & Largest $En$ \\
        \midrule

        Qwen3-VL-8B-Instruct & 1.20 & 0.92 & Top left, 6.64 \\
        InternVL3.5-8B & 1.18 & 0.95 & Bottom left, 2.29 \\
        GLM-4.1V-Thinking & 1.50 & 0.82 & Top right, 7.79 \\
        Nemotron-Omni & 1.43 & 0.87 & Top right, 3.86 \\
        \midrule
        Human boxes & 0.53 & 1.14 & Non-border, 1.87 \\
        \bottomrule
    \end{tabular}%
    }
\end{table}

%% file: tables/ablation_results2.tex
\begin{table}[t]
    \centering
    \caption{Performance across our 25-benchmark suite for the ablation study.}
    \label{tab:ablation_results}
    \setlength{\tabcolsep}{5pt}
    \renewcommand{\arraystretch}{1.15}

    \resizebox{0.85\columnwidth}{!}{%
    \begin{tabular}{lccccc|c}
        \toprule
        & \makecell{Qwen3-VL-8B\\Instruct}
        & coldstart
        & no\_sal
        & center\_rect
        & question\_boxes
        & \name{} \\
        \midrule
        \textbf{Mean score} ($\uparrow$) &
        61.14 & 62.69 & 63.47 & 63.39 & 63.33 & \textbf{64.26} \\
        \textbf{Mean rank} ($\downarrow$)
        & 3.90
        & 4.06
        & 3.34
        & 3.68
        & 3.38
        & \textbf{2.64} \\
        \textbf{Wins no.} ($\uparrow$)
        & 7
        & 0
        & 3
        & 2
        & 5
        & \textbf{9} \\
        \bottomrule
    \end{tabular}%
    }
\end{table}

%% file: sections/06_conclusion.tex
We introduced \name{}, a post-training method for vision-language models. Rather than prescribing target regions from the input image and question alone, \textsc{Self-Saliency} segments the model's reasoning chain into steps, grounds the objects mentioned in each observation step, and rewards the model for attending to those regions during GRPO training. We utilize a suite of 25 visual reasoning benchmarks and show that \textsc{Self-Saliency} achieves the highest mean score, best mean rank, and most wins, outperforming prior attention-steering methods. The gains are largest on tasks that depend on fine-grained visual evidence, and, unexpectedly, extend to mathematical reasoning benchmarks. After training, the model produces shorter reasoning chains which refer to larger image regions, and pays more attention to the grounded regions across relevant layers.

Additionally, we find a consistent geometric bias of attention towards the corners of the image, while human-annotated relevant regions lie mainly near the center. \textsc{Self-Saliency} substantially reduces this bias in selected heads. However, our ablations show that the gains cannot be explained by a simple attention alignment with the image center, nor by grounding question text alone. Grounding the model's own reasoning steps provides a more effective training signal.


%% file: sections/appendix_a.tex
\subsection{Reasoning Steps Segmentation and Classification} \label{app:steps_classifier}

Given a reasoning chain $c$ generated in response to a question $q$, we first decompose it into an ordered sequence of atomic reasoning steps,
$c = \langle s_1, \ldots, s_{|c|} \rangle$.
Each step is itself a sequence of tokens,
$s_i = \langle t_1, \ldots, t_{|s_i|} \rangle$,
and is assigned a functional label $l_i$.

We define four functional categories for reasoning steps: \emph{plan}, a forward-looking statement describing an intended action (e.g., ``let me check the left side of the image''); \emph{observe}, a direct statement about the visual content of the image (e.g., ``there is a red car on the left''); \emph{deduce}, a conclusion derived from previous reasoning steps rather than from newly introduced visual evidence (e.g., ``therefore it is a traffic sign''); and \emph{none}, a statement that does not contribute to the reasoning process (e.g., ``the user asked where the cat is'').

\paragraph{Segmentation.}
We segment the reasoning chain at sentence-final punctuation marks and line breaks, and discard fragments shorter than ten characters.

\paragraph{Classification.}
We classify each reasoning step using an encoder-only model trained by distilling labels produced by a proprietary model.\footnote{The resulting classifier achieves 91.9\% accuracy on the held-out test set and 93\% accuracy on a human-labeled test set.}

After classification, we retain only steps labeled as \emph{observe}, as these steps explicitly describe visual evidence from the image.

\subsection{Training}

\paragraph{Coldstart.}
For the coldstart, we use LoRA~\citep{hu2021lora} of rank 128 on all linear layers, learning rate $5\times10^{-5}$, and cosine schedule with $5\%$ warmup, training for two epochs.

\paragraph{GRPO training.}
We run GRPO~\citep{shao2024deepseekmath} for three epochs.
Each step draws $6$ prompts and samples $G=8$ rollouts per prompt at temperature $1.0$, for $48$ completions per step (one completion per device, gradient accumulation $8$). Prompts are truncated at $2{,}048$ tokens and completions at $1{,}024$, and images are resized so that the longer side is at most $512$ pixels. We update only LoRA adapters of rank $16$ ($\alpha=32$, dropout $0.05$) on the query and value projections, using AdamW at a learning rate of $1\times10^{-5}$ decayed linearly to zero with no warmup, no weight decay, and gradient-norm clipping at $1.0$. Advantages are normalized by the standard deviation of the group's rewards, and we use no KL penalty towards a reference policy ($\beta=0$), so the reward of Section~\ref{sec:rewards} is the entire training signal.

\paragraph{Judge model.}
As the LLM judge needed for $R_{llm}$ we use GPT-4o
mini~\citep{openai2024gpt4omini} prompted with the question, the ground-truth answer and the model's extracted answer span, and asked for a correctness rating on a $1$--$5$ scale that we rescale to $[0,1]$.

\subsection{Evaluation}

\textbf{Evaluation protocol.}
We evaluate with lmms-eval~\citep{zhang2025lmms}, temperature $0$, single beam with
\texttt{max\_new\_tokens}$=4096$, \texttt{repetition\_penalty}$=1.05$, batch size
$1$, in bfloat16.
Images are processed at lmms-eval's default budget of $1{,}605{,}632$ pixels, raised to $3{,}211{,}264$ for the three benchmarks whose subject is resolution (MME-RealWorld, HR-Bench 4K/8K). Benchmarks whose scoring requires answer extraction or an LLM judge use GPT-4o mini~\citep{openai2024gpt4omini}.

\textbf{Benchmark splits.}
We use the following splits for the evaluation benchmarks: \emph{soft\_localization} for IllusionVQA, \emph{testmini\_vision\_only} for MathVerse, \emph{testmini} for MathVision, \emph{testmini} for MathVista, \emph{standard} for MMMUPro, \emph{test} for Omnispatial, \emph{p3} for SalBench, \emph{img} for ScienceQA, \emph{testmini} for WeMath.

%% file: sections/appendix_b.tex
In this section, we report the results over our 25-benchmark suite, with bootstrap standard errors.

\paragraph{Error bars.}
Each reported score carries a nonparametric bootstrap estimate of its question-sampling uncertainty: we resample the run's units with replacement 10,000 times, recompute the score, and report the standard deviation of the resampled scores. The unit is whatever the benchmark's aggregation treats as independent -- the question for most tasks, the image for MME (which asks two questions per image), and the question rather than the row for HR-Bench (where each is asked four times under cyclic option rotation); groups are resampled independently to their original sizes, so that macro-averaged metrics such as MMStar's nested average are preserved. Differences between models use a paired bootstrap over the questions both models answered, which is markedly tighter than combining the two marginal bars, since questions the models score identically contribute no uncertainty.

\input{tables/results_with_errors}

\paragraph{Results.}
We show results in Table~\ref{tab:results_with_errors}.

%% file: tables/results_with_errors.tex
\providecommand{\name}{Self-Saliency}  

\begin{table*}[t]
\centering
\newcommand{\Vstd}[2]{#1\,{\tiny$\pm$#2}}           
\newcommand{\Bstd}[2]{\textbf{#1}\,{\tiny$\pm$#2}}  
\caption{Results across 25 benchmarks (mean $\pm$ standard error). Best per benchmark in \textbf{bold}. Mean score is computed over all 25 benchmarks after rescaling MME to $[0,100]$ (raw score $/\,2800$). Mean rank is the average over per-benchmark ranks (1 = best).}
\label{tab:results_with_errors}
\resizebox{\textwidth}{!}{%
\begin{tabular}{l ccc ccc c}
\toprule
& \multicolumn{3}{c}{\textit{No attention intervention}} & \multicolumn{3}{c}{\textit{Prior work}} & \multicolumn{1}{c}{\textit{Ours}} \\
\cmidrule(lr){2-4} \cmidrule(lr){5-7} \cmidrule(lr){8-8}
Benchmark & \makecell{Qwen3-VL\\8B-Instruct} & Coldstart & No-Sal & VGA & \makecell{Saliency-\\R1} & EASE & \name{} \\
\midrule
AlgoPuzzleVQA $\uparrow$  & \Vstd{32.8}{1.11} & \Vstd{36.9}{1.14} & \Vstd{39.9}{1.15} & \Vstd{34.3}{1.12} & \Vstd{38.7}{1.15} & \Vstd{36.4}{1.13} & \Bstd{41.4}{1.16} \\
ChartQA $\uparrow$        & \Vstd{85.0}{0.71} & \Vstd{88.8}{0.63} & \Vstd{89.4}{0.62} & \Vstd{84.9}{0.72} & \Vstd{89.7}{0.61} & \Vstd{88.9}{0.63} & \Bstd{90.0}{0.60} \\
CV-Bench $\uparrow$       & \Vstd{86.0}{0.68} & \Vstd{83.4}{0.72} & \Vstd{84.9}{0.70} & \Bstd{86.1}{0.67} & \Vstd{83.8}{0.72} & \Vstd{84.2}{0.71} & \Vstd{85.9}{0.68} \\
DailyClue $\uparrow$      & \Bstd{35.4}{1.85} & \Vstd{32.4}{1.82} & \Vstd{34.2}{1.84} & \Vstd{34.2}{1.84} & \Vstd{32.0}{1.81} & \Vstd{32.7}{1.82} & \Vstd{31.2}{1.80} \\
HallusionBench $\uparrow$ & \Bstd{63.2}{1.56} & \Vstd{59.7}{1.59} & \Vstd{57.1}{1.61} & \Vstd{62.0}{1.57} & \Vstd{58.3}{1.60} & \Vstd{62.6}{1.57} & \Vstd{55.7}{1.57} \\ 
HRBench4k $\uparrow$      & \Vstd{78.0}{2.70} & \Vstd{76.9}{2.70} & \Vstd{77.4}{2.63} & \Vstd{78.1}{2.68} & \Vstd{78.9}{2.57} & \Vstd{76.9}{2.69} & \Bstd{79.5}{2.60} \\
HRBench8k $\uparrow$      & \Vstd{74.1}{2.75} & \Vstd{72.9}{2.77} & \Vstd{72.0}{2.79} & \Vstd{74.2}{2.77} & \Bstd{76.0}{2.66} & \Vstd{72.8}{2.86} & \Vstd{74.5}{2.73} \\
IllusionVQA $\uparrow$    & \Vstd{40.0}{1.55} & \Vstd{41.9}{1.56} & \Vstd{41.5}{1.56} & \Vstd{40.1}{1.55} & \Bstd{43.8}{1.57} & \Vstd{40.5}{1.55} & \Vstd{43.5}{1.57} \\
LogicVista $\uparrow$     & \Vstd{52.5}{2.36} & \Vstd{54.9}{2.35} & \Vstd{55.1}{2.35} & \Vstd{51.3}{2.36} & \Bstd{56.5}{2.35} & \Vstd{52.9}{2.36} & \Vstd{56.2}{2.35} \\
MathVerse $\uparrow$      & \Vstd{34.4}{1.69} & \Vstd{51.0}{1.78} & \Bstd{52.8}{1.78} & \Vstd{33.8}{1.69} & \Vstd{49.5}{1.78} & \Vstd{50.4}{1.78} & \Vstd{50.9}{1.78} \\
MathVision $\uparrow$     & \Vstd{29.6}{2.62} & \Vstd{33.5}{2.71} & \Vstd{37.5}{2.78} & \Vstd{31.2}{2.66} & \Vstd{33.9}{2.72} & \Vstd{35.9}{2.76} & \Bstd{41.5}{2.83} \\
MathVista $\uparrow$      & \Vstd{73.5}{1.40} & \Vstd{69.5}{1.46} & \Vstd{71.8}{1.42} & \Bstd{73.8}{1.39} & \Vstd{71.0}{1.44} & \Vstd{70.0}{1.45} & \Vstd{70.4}{1.44} \\
MME $\uparrow$            & \Vstd{2375.9}{36.2} & \Vstd{2371.7}{35.1} & \Vstd{2376.1}{36.3} & \Vstd{2379.8}{36.2} & \Bstd{2395.7}{34.0} & \Vstd{2395.4}{33.1} & \Vstd{2355.1}{35.1} \\
MME-RW $\uparrow$         & \Vstd{63.7}{0.31} & \Vstd{64.2}{0.31} & \Vstd{66.7}{0.31} & \Vstd{63.8}{0.31} & \Vstd{66.4}{0.31} & \Vstd{65.1}{0.31} & \Bstd{68.7}{0.30} \\
MMK12 $\uparrow$          & \Bstd{68.2}{1.04} & \Vstd{63.1}{1.08} & \Vstd{62.3}{1.08} & \Vstd{67.9}{1.04} & \Vstd{61.1}{1.09} & \Vstd{63.1}{1.08} & \Vstd{62.0}{1.09} \\
MMMUPro $\uparrow$        & \Vstd{40.4}{1.18} & \Vstd{48.6}{1.20} & \Bstd{48.6}{1.20} & \Vstd{41.9}{1.19} & \Vstd{47.7}{1.20} & \Vstd{47.1}{1.20} & \Vstd{48.4}{1.20} \\
MMStar $\uparrow$         & \Vstd{64.0}{1.34} & \Vstd{69.8}{1.29} & \Bstd{70.9}{1.26} & \Vstd{64.7}{1.33} & \Vstd{69.9}{1.28} & \Vstd{69.5}{1.28} & \Vstd{69.6}{1.28} \\
Omnispatial $\uparrow$    & \Vstd{47.4}{1.28} & \Vstd{50.3}{1.28} & \Vstd{48.2}{1.28} & \Vstd{46.8}{1.27} & \Vstd{50.2}{1.28} & \Vstd{50.2}{1.28} & \Bstd{51.0}{1.28} \\
SalBench $\uparrow$       & \Vstd{43.3}{0.97} & \Vstd{60.7}{0.96} & \Vstd{59.8}{0.96} & \Vstd{43.3}{0.97} & \Vstd{62.7}{0.95} & \Vstd{58.6}{0.97} & \Bstd{64.7}{0.94} \\
POPE $\uparrow$           & \Vstd{88.8}{0.33} & \Vstd{87.4}{0.35} & \Vstd{88.2}{0.34} & \Vstd{88.8}{0.33} & \Vstd{88.5}{0.34} & \Vstd{88.5}{0.34} & \Bstd{88.9}{0.33} \\
RealWorldQA $\uparrow$    & \Vstd{68.5}{1.68} & \Vstd{67.1}{1.70} & \Vstd{67.8}{1.69} & \Vstd{68.4}{1.68} & \Bstd{68.8}{1.68} & \Vstd{67.5}{1.70} & \Vstd{68.2}{1.68} \\
ScienceQA $\uparrow$      & \Vstd{94.4}{0.51} & \Vstd{96.3}{0.42} & \Vstd{97.2}{0.37} & \Vstd{94.4}{0.51} & \Bstd{97.6}{0.34} & \Vstd{96.2}{0.42} & \Vstd{97.4}{0.35} \\
VisuLogic $\uparrow$      & \Bstd{26.6}{1.40} & \Vstd{24.9}{1.37} & \Vstd{24.8}{1.37} & \Vstd{26.4}{1.39} & \Vstd{24.9}{1.37} & \Vstd{24.4}{1.36} & \Vstd{24.4}{1.36} \\
V$^{*}$ $\uparrow$        & \Vstd{84.3}{2.64} & \Vstd{79.6}{2.92} & \Vstd{79.6}{2.92} & \Bstd{85.3}{2.57} & \Vstd{80.6}{2.87} & \Vstd{81.2}{2.84} & \Vstd{82.7}{2.74} \\
WeMath $\uparrow$         & \Vstd{69.4}{1.10} & \Vstd{68.6}{1.11} & \Vstd{74.3}{1.05} & \Vstd{68.2}{1.12} & \Vstd{73.8}{1.05} & \Vstd{71.1}{1.09} & \Bstd{75.5}{1.03} \\
\midrule
Mean score $\uparrow$     & 61.1 & 62.7 & 63.5 & 61.2 & 63.6 & 62.9 & \textbf{64.3} \\
Mean rank $\downarrow$    & 4.42 & 4.64 & 3.68 & 4.48 & 3.26 & 4.50 & \textbf{3.02} \\
Number of wins $\uparrow$ & 4 & 0 & 3 & 3 & 6 & 0 & \textbf{9} \\
\bottomrule
\end{tabular}}
\end{table*}

%% file: sections/appendix_c.tex
\newcommand{\namemd}[0]{\name{}$_{md}$}

\paragraph{Definition.}
We define a variant of our saliency score $\phi$ (Equation~\ref{eq:phi}) in which the denominator is the mean saliency over the image rather than the maximum saliency:

\begin{equation*}
\phi_{\mathrm{mean}}(s)
:=
\frac{
\frac{1}{|u_s|}
\sum_{p \in u_s} sal_s(p)
}{
\frac{1}{|I|} \sum_{p \in I} sal_s(p)
}.
\end{equation*}

Analogously to the saliency reward, we define the corresponding mean-denominator saliency reward as

\begin{equation*}
R_{sal\_\mathrm{mean}} =
\frac{1}{|O|}
\sum_{s \in O} \phi_{\mathrm{mean}}(s)
\end{equation*}

\paragraph{Training.}
We follow the same training procedure described in Section~\ref{sec:training}, while calibrating the reward weight $\alpha_{sal\_\mathrm{mean}}$ relative to the tuned value of $\alpha_{sal}$. To approximately match the strength of the training signal applied by GRPO, we measure the within-group standard deviation of $R_{sal}$ and $R_{sal\_\mathrm{mean}}$ over groups of completions sampled from the cold-start model, and set

\begin{equation*}
\alpha_{sal\_\mathrm{mean}}
=
\alpha_{sal}
\times
\frac{\operatorname{std}(R_{sal})}
{\operatorname{std}(R_{sal\_\mathrm{mean}})}.
\end{equation*}

Given our tuned value of $\alpha_{sal}=0.4$, this yields $\alpha_{sal\_\mathrm{mean}}=0.033$.
We denote the resulting checkpoint by \namemean{}.

\paragraph{Evaluation.}
We evaluate \namemean{} on the same suite of 25 benchmarks used in Section~\ref{sec:evaluation}.

\input{tables/results_with_md}

\paragraph{Results.}
Results are presented in Table~\ref{tab:results_with_md}. Unlike \name{}, \namemean{} does not improve over the \emph{no\_sal} baseline, which is trained without attention steering.

%% file: tables/results_with_md.tex
\providecommand{\name}{Self-Saliency}

\begin{table*}[t]
\centering
\caption{Results across 25 benchmarks, including \namemean{}. Best per benchmark in \textbf{bold}.}
\label{tab:results_with_md}
\resizebox{\textwidth}{!}{%
\begin{tabular}{ll ccc cc}
\toprule
& & \multicolumn{3}{c}{\textit{No attention intervention}} & \multicolumn{2}{c}{\textit{Saliency-based rewards}} \\
\cmidrule(lr){3-5} \cmidrule(lr){6-7}
Benchmark & Split & \makecell{Qwen3-VL\\8B-Instruct} & Coldstart & No-Sal & \namemean{} & \name{} \\
\midrule
AlgoPuzzleVQA $\uparrow$ & -- & 32.8 & 36.9 & 39.9 & 38.8 & \textbf{41.4} \\
ChartQA $\uparrow$ & -- & 85.0 & 88.8 & 89.4 & 89.7 & \textbf{90.0} \\
CV-Bench $\uparrow$ & -- & \textbf{86.0} & 83.4 & 84.9 & 84.4 & 85.9 \\
DailyClue $\uparrow$ & -- & \textbf{35.4} & 32.4 & 34.2 & 32.7 & 31.2 \\
HallusionBench $\uparrow$ & image & \textbf{63.2} & 59.7 & 57.1 & 55.7 & 55.7 \\
HRBench $\uparrow$ & 4k & 78.0 & 76.9 & 77.4 & 76.8 & \textbf{79.5} \\
HRBench $\uparrow$ & 8k & 74.1 & 72.9 & 72.0 & \textbf{74.5} & \textbf{74.5} \\
IllusionVQA $\uparrow$ & soft\_localization & 40.0 & 41.9 & 41.5 & \textbf{44.1} & 43.5 \\
LogicVista $\uparrow$ & -- & 52.5 & 54.9 & 55.1 & 53.1 & \textbf{56.2} \\
MathVerse $\uparrow$ & testmini\_vision\_only & 34.4 & 51.0 & \textbf{52.8} & 50.8 & 50.9 \\
MathVision $\uparrow$ & testmini & 29.6 & 33.5 & 37.5 & 36.8 & \textbf{41.5} \\
MathVista $\uparrow$ & testmini & \textbf{73.5} & 69.5 & 71.8 & 70.8 & 70.4 \\
MME $\uparrow$ & -- & 2375.9 & 2371.7 & \textbf{2376.1} & 2359.4 & 2355.1 \\
MME-RW $\uparrow$ & -- & 63.7 & 64.2 & 66.7 & 66.6 & \textbf{68.7} \\
MMK12 $\uparrow$ & -- & \textbf{68.2} & 63.1 & 62.3 & 60.2 & 62.0 \\
MMMUPro $\uparrow$ & standard & 40.4 & 48.6 & \textbf{48.6} & 48.3 & 48.4 \\
MMStar $\uparrow$ & -- & 64.0 & 69.8 & \textbf{70.9} & 70.3 & 69.6 \\
Omnispatial $\uparrow$ & test & 47.4 & 50.3 & 48.2 & 50.0 & \textbf{51.0} \\
SalBench $\uparrow$ & p3 & 43.3 & 60.7 & 59.8 & 61.9 & \textbf{64.7} \\
POPE $\uparrow$ & -- & 88.8 & 87.4 & 88.2 & 88.2 & \textbf{88.9} \\
RealWorldQA $\uparrow$ & -- & \textbf{68.5} & 67.1 & 67.8 & 68.2 & 68.2 \\
ScienceQA $\uparrow$ & img & 94.4 & 96.3 & 97.2 & 96.9 & \textbf{97.4} \\
VisuLogic $\uparrow$ & -- & \textbf{26.6} & 24.9 & 24.8 & 23.3 & 24.4 \\
V$^{*}$ $\uparrow$ & -- & \textbf{84.3} & 79.6 & 79.6 & 79.6 & 82.7 \\
WeMath $\uparrow$ & testmini & 69.4 & 68.6 & 74.3 & 73.4 & \textbf{75.5} \\
\midrule
Mean score $\uparrow$ & -- & 61.1 & 62.7 & 63.5 & 63.2 & \textbf{64.3} \\
Mean rank $\downarrow$ & -- & 3.24 & 3.52 & 2.68 & 3.26 & \textbf{2.30} \\
\bottomrule
\end{tabular}}
\end{table*}

%% file: sections/appendix_d.tex
In this section, we describe our reimplementation of baselines without a published checkpoint.

\subsection{VGA}
VGA~\citep{zhao2026tell} is training-free, so our replication is an inference-time patch
on Qwen3-VL rather than a retrained checkpoint. We keep the
published hyperparameters -- the guidance strength $\beta$, the top-$K$ of the
object-agnostic map, and the head-balancing scheme. Two points still required a decision, in each case
because the original is backbone-specific or under-specified. We flag them
in the following paragraphs.

\emph{The layer window.} The window over which guidance is applied is
backbone-specific in the original, and the rule for choosing it -- start where
attention to the sink position begins to dominate -- fixes neither a quantity
nor a threshold. On Qwen3-VL-8B its two natural readings disagree, selecting
layer~$3$ or layer~$7$ against the published~$4$. Rather than choose, we swept
five windows spanning both ($\mathrm{start} \in \{2,4,7\}$,
$\mathrm{end} \in \{16,20\}$, $\beta \in \{0.2,0.25\}$) over $13$ benchmarks at
$100$ documents each. The published window scores highest and the spread across
all five is $0.005$, so we report VGA as published; the result is not an artefact
of under-tuning.

\emph{The object's first token.} The map reads the probability of the object's
first token at each visual position, but under BPE a word and its
space-prefixed form are distinct tokens and it is not stated which is intended.
We score the plain, space-prefixed and capitalised forms and keep the largest,
which can only raise a real object above the floor.

\subsection{Saliency-R1}
We reimplement Saliency-R1 \citep{gong2026saliency} from the authors' public
repository\footnote{\url{https://github.com/peterant330/Saliency_R1}}, changing as little as the backbone swap allows. Their released
checkpoints are Qwen2.5-VL-3B and -7B; no Qwen3-VL model is released.

\paragraph{Cold start.} The cold-start stage described in Section~\ref{sec:training}
\emph{is} theirs, simply on Qwen3-VL-8B-Instruct instead of Qwen2.5-VL-7B-Instruct. The
Saliency-R1 row and all our own rows therefore begin from the same weights, and
the comparison isolates the RL stage.

\paragraph{Reward.} Their saliency map is computed as released: for each rollout,
the attention of the answer tokens to the reasoning span is composed with the
reasoning span's attention to the image tokens, weighted by the corresponding
value vectors, projected through the output projection and the unembedding,
summed over answer tokens, rectified, and reshaped onto the merged patch grid.
The reward is the fraction of that map's mass falling inside the question-level
box shipped with Saliency-R1-8K, and zero for completions that fail their format
check. Our only edit to this code is that rather than retaining per-step attention throughout
sampling, we sample first and recover the same attention rows from a single
teacher-forced pass over the sampled sequence. Under causal attention the two are algebraically identical, and the second fits 1{,}024-token completions in memory.

\paragraph{GRPO configuration.} We keep their published command: the same corpus
and the same 100-sample held-out split (seed 42), their system prompt, learning
rate $1\times10^{-5}$, three epochs, $G=8$ rollouts at temperature $1.0$, LoRA on
the query and value projections at the library defaults their command leaves
unset (rank 16, $\alpha=32$, dropout $0.05$), no KL penalty, and their global
batch of 64 completions per optimizer step -- redistributed as one completion per
device with gradient accumulation 8 over eight GPUs, rather than their four and
two, for memory. All eight GPUs
train, giving 8 prompts per step and $2{,}991$ optimizer steps for three epochs
rather than the $3{,}990$ of Section~\ref{sec:training}. We deviate from their
command in three places. First, prompts and completions are capped at $2{,}048$
and $1{,}024$ tokens instead of $1{,}024$ and $512$, matching our other runs; the
cap was rarely active, with mean completion length between 162 and 247 tokens and
fewer than $0.5\%$ of rollouts reaching it. Second, we add the same exact-match
reward $R_{\text{direct}}$ that every other arm receives, so the reward here is
$R_{\text{format}}+R_{\text{direct}}+R_{\text{llm}}+R^{\text{S-R1}}_{\text{sal}}$
and this arm differs from ours in the attention term alone; we leave their weight
of $1$ on $R^{\text{S-R1}}_{\text{sal}}$ rather than retuning it to our
$\alpha_{\text{sal}}$. Third, training-time sampling uses no repetition penalty,
where their command sets $1.05$. We use the same correctness judge they use, GPT-4o-mini with their prompt.

\subsection{EASE}

Although EASE reports results on Qwen3-VL-8B, the authors did not release their evaluation protocol, training data, or a trained checkpoint. We contacted the authors by email but did not receive a response. We therefore reimplemented and trained EASE ourselves and evaluated it using our unified evaluation protocol.

\paragraph{What we run, and on what data.} We use their code as released, in their
framework, in a separate environment; the files implementing the method itself are
untouched, and our changes are confined to their reward interface and an import
fix. They release neither the evidence-box annotations nor the code of the three-stage
pipeline that produced them, which is described in prose only. We therefore run
EASE on Saliency-R1-8K, which already ships a box for every one of its $8{,}080$
rows, bypassing their annotation pipeline entirely. This has a second benefit:
EASE and our method then differ in exactly one thing -- where the box signal
enters -- on the same corpus, from the same cold-start checkpoint, with the same
held-out split and the same judge.

\paragraph{Verification.} Because EASE is a method inside a framework that is not
ours, a single EASE number would confound the two. We therefore also trained their
DAPO baseline in the same framework, on the same data, from the same cold start.
On the natural-image benchmarks where both arms were scored, EASE $-$ DAPO is
$+2.52$ points, against the $+2.5$ to $+3.1$ over DAPO that they report: the
effect replicates, on our corpus, in their framework. One qualification belongs
with that number -- DAPO is a weak reference here, scoring slightly below the
cold start it started from, so part of the margin is EASE avoiding damage rather
than adding value; measured against the cold start directly, EASE is $+2.07$.

%% file: sections/appendix_e.tex
Table~\ref{tab:attention_exp_models} shows the characteristics of each of the models we use in the experiment in Section~\ref{sec:analysis}.

\input{tables/attention_exp_models}

%% file: tables/attention_exp_models.tex
\begin{table}[t]
    \centering
    \caption{Models used in the attention focus experiment.}
    \label{tab:attention_exp_models}

    \resizebox{0.8\columnwidth}{!}{%
    \begin{tabular}{lccc}
        \toprule
        Model & Vision encoder & Grid shape & Language model \\
        \midrule

        Qwen3-VL-8B-Instruct & ViT & Image-dependent & Qwen3-8B \\
        InternVL3.5-8B & InternViT-300M & 16×16 & Qwen3-8B \\
        GLM-4.1V-Thinking & AIMv2-Huge & Image-dependent & GLM-4-9B-0414 \\
        Nemotron-Omni & C-RADIOv4-H & Image-dependent & Nemotron \\

        \bottomrule
    \end{tabular}%
    }
\end{table}